\documentclass[runningheads]{llncs}
\usepackage{xcolor}
\usepackage[misc]{ifsym}
\usepackage{graphicx}
\usepackage{booktabs}   
\usepackage{multirow}   
\usepackage{amsmath}
\usepackage{amssymb}
\usepackage[colorlinks=true, urlcolor=blue, linkcolor=red]{hyperref}

\usepackage{booktabs}
\usepackage{cite}
\usepackage{bbding}
\usepackage{xcolor}
\usepackage{makecell}

\begin{document}
\title{Learning Directional Semantic Transitions for Longitudinal Chest X-ray Analysis}
\titlerunning{Semantic Transitions for Longitudinal CXR Analysis}
\author{
Zhangfeng Hu\inst{1}
\and
Zefan Yang\inst{1}
\and
Ge Wang\inst{1}
\and
Tanveer Syeda-Mahmood\inst{2}
\and
Anushree Burade\inst{3}
\and
Mannudeep Kalra\inst{3}
\and
Pingkun Yan\inst{1}\Letter
}

\authorrunning{Hu et al.}

\institute{
Rensselaer Polytechnic Institute, Troy, NY, USA\\
\email{yanp2@rpi.edu}
\and
Stanford University, Stanford, CA, USA\\
\and
Massachusetts General Hospital, Harvard Medical School, Boston, MA, USA
}

\maketitle              

\begin{abstract}
 Chest X-ray (CXR) interpretation often requires longitudinal comparison to assess disease progression. Existing approaches typically rely on temporal feature fusion or inter-study discrepancy modeling, yet remain limited in capturing subtle progression semantics and overlook the inherently directional nature of disease trajectories. In this paper, we propose ProTrans, a novel vision-language pretraining framework that formulates disease progression as a directional semantic transition between paired CXR studies. ProTrans leverages radiology reports to anchor individual CXR representations within interpretable disease states, and introduces a learnable progression feature map to explicitly encode semantic shifts between states, aligned with report-derived progression descriptions. To enforce direction-aware perception, ProTrans incorporates a reversed temporal modeling process and imposes bidirectional reconstruction consistency across states and transitions, thereby disentangling directional semantics and promoting coherent trajectory modeling. Extensive experiments on longitudinal downstream tasks, including disease progression classification and progression captioning, demonstrate that ProTrans consistently outperforms existing methods, establishing a unified pretraining framework for longitudinal CXR understanding. \url{https://github.com/RPIDIAL/ProTrans}
\end{abstract}

\section{Introduction}
Recent advances in vision–language pretraining (VLP) have substantially improved chest X-ray (CXR) analysis, demonstrating superior performance in tasks such as abnormality localization, classification, and report generation \cite{huang2021gloria, boecking2022making, wang2022multi}. By aligning visual representations with clinical concepts derived from paired radiology reports, these models capture rich semantic information essential for CXR interpretation. Despite these successes, existing approaches primarily focus on single-study analysis. In clinical practice, however, radiologists routinely compare the current CXR with prior examinations to identify subtle temporal changes and assess disease progression \cite{zhou2025review}. Such longitudinal assessment is pivotal to diagnosis, treatment, and prognosis. 

To bridge this gap, recent efforts have begun extending vision architectures to capture longitudinal information, which can be broadly categorized into two groups. The first group includes temporal fusion modules to aggregate multi-time-point features, including cross-attention encoders \cite{bannur2023learning,chen2025coca,yang2025tempa,liu2025priorrg} to exploit temporal correlations, 
temporal alignment connectors to integrate multi-scale context from priors \cite{zhang2025libra}, and group causal transformers to model temporally ordered dependencies in long sequences \cite{wang2024hergen}. The second category models inter-study discrepancies by extracting patch-, region-, or multi-level feature differences \cite{yun2025diff,wang2025chexlearner,song2025ddatr}. Despite promising results, it remains unclear whether these methods can accurately and reliably distill clinically meaningful progression semantics. Two key challenges persist. First, consecutive CXRs are dominated by invariant anatomical structures, where true progression cues are subtle and easily overwhelmed. Meanwhile, pseudo-changes caused by inconsistent imaging conditions can confound both naive feature fusion and differentiation modeling. Second, disease progression is inherently directional. In addition to change detection, clinically useful models should characterize their trajectories (\textit{e.g.}, improving versus worsening). However, this directional nature of progression remains largely underexplored in existing approaches.

To address these limitations, we propose \textbf{ProTrans}, a novel VLP framework for progression representation learning that models disease \textbf{Pro}gression as directional semantic \textbf{Trans}itions between consecutive CXR studies. Rather than capturing temporal change through feature-level fusion or subtraction, ProTrans formulates progression as a learnable transformation in a semantic space. Specifically, we introduce a learnable progression feature map that resides between two longitudinal CXRs, as a bridge to encode the semantic transition between prior and current states. To achieve this goal, we first design a state contrastive alignment strategy that leverages associated radiology reports to ground each CXR representation in a report-based semantic state. Then, a transition contrastive constraint aligns progression representations with disease transition cues derived from paired reports, encouraging them to capture clinical state changes. To further enforce temporal directionality, ProTrans introduces a reversed temporal learning process to estimate reversed progression representations. Then we impose a bidirectional reconstruction consistency ensuring that the current state can be reconstructed from the prior state and progression, and vice versa using the reversed progression. By jointly learning forward and backward transitions, the model disentangles directional temporal semantics and improves sensitivity to trajectory-specific disease evolution. Together, these components enable a unified pretraining framework for semantically-sensitive and direction-aware progression representation learning. 

\begin{figure}[t]
    \centering
\includegraphics[width=0.75\linewidth]{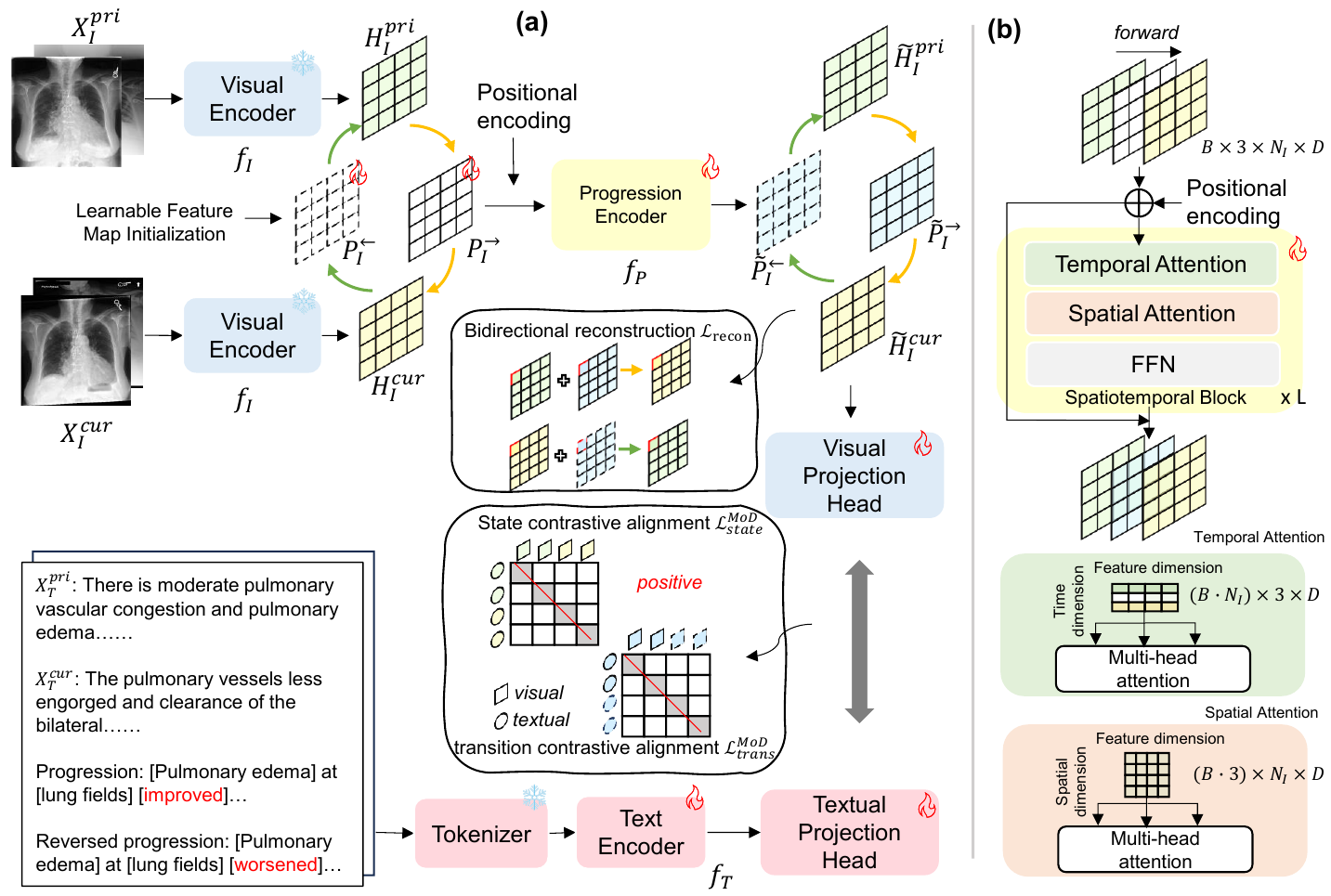}
\vspace{-5pt}
    \caption{\textbf{Overview of ProTrans.} (a) Pretraining pipeline for progression representation learning, and (b) Architecture of the progression encoder.}
    \label{fig:framework}
\end{figure}

Our contributions are three-fold. \textbf{(1)} We introduce ProTrans, a novel progression representation learning framework that formulates disease progression as directional semantic transitions, enhanced by reversed temporal modeling and bidirectional reconstruction consistency to capture a coherent disease trajectory. \textbf{(2)} We propose joint state and transition contrastive alignment that grounds bi-temporal clinical states and their interval transitions within a shared semantic space based on radiology reports, enabling well-grounded and interpretable encoding of disease evolution. \textbf{(3)} We demonstrate that ProTrans improves performance across longitudinal downstream tasks, including disease progression classification and progression captioning, providing a unified pretraining foundation for longitudinal CXR understanding.

\section{Method}

Fig.\ref{fig:framework}(a) illustrates the framework of ProTrans with details showing how disease progression is modeled as a directional semantic transition between longitudinal CXR studies. The framework learns state representations and progression transformations jointly, integrating bidirectional temporal modeling and reconstruction constraints to enforce trajectory-aware representation learning.

{\textbf{Bidirectional Progression Representation Modeling}.} Given a pair of prior and current CXR images $X_I^{pri}$, $X_I^{cur}$ and their corresponding reports $X_T^{pri}$, $X_T^{cur}$, we first use visual encoder $f_I$ and text encoder $f_T$ to extract visual and textual embeddings as $\mathbf{H}_I^{t}\in\mathbb{R}^{N_I\times d}$, $\mathbf{H}_T^{t}\in\mathbb{R}^{N_T\times d}$, where $t\in\{\mathrm{pri},\mathrm{cur}\}$. $N_I$, $N_T$ denote the numbers of visual and textual tokens, and $d$ is the token dimension. 

Then to explicitly model progression semantics, we introduce a learnable progression feature map $\mathbf{P}_I^{\rightarrow} \in \mathbb{R}^{N_I \times d}$, which is inserted between prior and current image embeddings to form a directed temporal sequence as $\mathbf{S}^{\rightarrow} = [\mathbf{H}_I^{pri}; \mathbf{P}_I^{\rightarrow}; \mathbf{H}_I^{cur}] \\ \in\mathbb{R}^{3 \times N_I\times d}$. Positional encoding is integrated to preserve the ordering information within the sequence. The sequence is then processed by a progression encoder $f_P$, as depicted in Fig.\ref{fig:framework}(b), composed of $L$ spatiotemporal blocks. Each block contains a temporal attention layer to model cross-time dependencies, a spatial attention layer to capture anatomical correspondence, and a feed-forward network (FFN), all equipped with residual connections. After processing by $f_P$, we obtain a set of progression-aware representations as $[\tilde{\mathbf{H}}_I^{pri}, \; \tilde{\mathbf{P}}_I^{\rightarrow}, \; \tilde{\mathbf{H}}_I^{cur}] = f_P(\mathbf{S}^{\rightarrow})
$. To enhance sensitivity to temporal directionality, we model the reversed progression by constructing a backward sequence as $\mathbf{S}^{\leftarrow} = [\mathbf{H}_I^{cur}; \mathbf{P}_I^{\leftarrow}; \mathbf{H}_I^{pri}]$, where $\mathbf{P}_I^{\leftarrow}$ is a learnable reversed progression map and produces the corresponding reversed progression representation $\tilde{\mathbf{P}}_I^{\leftarrow}$.

\textbf{{State Contrastive Alignment}.} To facilitate above progression representation learning, we devise joint state-transition contrastive alignment to ground bi-temporal clinical  states and their interval transitions within a shared semantic space based on radiology reports. First, to ensure that each CXR representation corresponds to an interpretable clinical state, we perform cross-modal contrastive alignment between images and their paired reports at each time point. Specifically, given a current image $\tilde{\mathbf{H}}_{I,i}^{cur}$ of sample $i$, its associated report $\mathbf{H}_{T,i}^{cur}$ is treated as the positive, while all other reports in the mini-batch $B$—including those from different samples and the prior report of sample $i$—are treated as negatives. This design enforces fine-grained temporal discrimination, encouraging the model to distinguish subtle semantic differences across longitudinal states. The same alignment is applied to the prior study. Formally, let $\mathbf{h}_{I,i}^{t}$ and $\mathbf{h}_{T,i}^{t}$ denote the normalized and projected embeddings from $\tilde{\mathbf{H}}_{I,i}^{t}$, $\mathbf{H}_{T,i}^{t}$. The image-to-text similarity logits $p_{I2T}^t \in \mathbb{R}^{B \times B}$ are computed as:

{\small
\begin{equation}
p_{I2T,i}^t =
\frac{\exp(\mathrm{sim}(\mathbf{h}_{I,i}^{t}, \mathbf{h}_{T,i}^{t}) / \tau)}{\sum_{k=1}^{B} [\exp(\mathrm{sim}(\mathbf{h}_{I,i}^{t}\mathbf{h}_{T,k}^{t}) / \tau) + \exp(\mathrm{sim}(\mathbf{h}_{I,i}^{t}, \mathbf{h}_{T,k}^{\bar{t}}) / \tau)]},
\label{logit}
\end{equation}}
where $\tau$ is a temperature parameter and $\bar{t}$ denotes the complementary time-point (i.e., $\bar{\mathrm{pri}}=\mathrm{cur}$ and $\bar{\mathrm{cur}}=\mathrm{pri}$), $\mathrm{sim}(\cdot,\cdot)$ denotes a token-wise similarity function  \cite{yao2021filip}. Similarly, the text-to-image similarity logits are described as $p_{T2I}^t$. Then, the cross-modal alignment loss is defined as:

{\small
\begin{equation}
\mathcal{L}_{\mathrm{state}} = -\frac{1}{2B} \sum_{t \in \{pri,cur\}} \sum_{k=1}^{B} (q^t_k \log p_{I2T,k}^t + q_k^t \log p_{T2I,k}^t), 
\label{nce}
\end{equation}}
where $q^t$ is the one-hot alignment label matrix. 
Since longitudinal CXR studies often exhibit semantic overlap, such strict one-hot alignment may be overly restrictive. We therefore introduce additional soft targets \cite{li2021align}, generated by a momentum model—an exponential-moving-average (EMA) version of the base network. Specifically, momentum image-text similarity logits $q'^t_{I2T}$ and $q'^t_{T2I}$ are computed using momentum embeddings $\mathbf{h}'$ replacing $\mathbf{h}$ in Eq.(\ref{logit}), and then serve as soft supervisory signals replacing $q^t$ in Eq.(\ref{nce}) to obtain a soft alignment loss $\mathcal{L}^{soft}_{\mathrm{state}}$. The final state alignment loss is defined as:
{\small
\begin{equation}
\mathcal{L}_{\mathrm{state}}^{\mathrm{MoD}}
= \alpha\mathcal{L}_{\mathrm{state}} + (1-\alpha)\mathcal{L}^{soft}_{\mathrm{state}},
\label{state}
\end{equation}}

where $\alpha$ controls the trade-off between hard and soft alignment supervision.

\textbf{Transition Contrastive Alignment.} To ensure that learned progression representations $\tilde{\mathbf{P}}_{I}^{\rightarrow}$, $\tilde{\mathbf{P}}_{I}^{\leftarrow}$ capture meaningful transition semantics, we further align them with corresponding textual descriptions of disease progression. Specifically, we construct structured transition texts in the form of ``[finding] at [position] [progression]'', based on the disease progression annotations from Chest ImaGenome dataset \cite{wu2021chest}. For reversed transitions, we simply use opposite phrasing (e.g., replacing ``improved'' with ``worsened'', while ``no change'' remains unchanged), thereby encoding direction-aware semantics. These structured descriptions are encoded by the text encoder $f_T$ into transition embeddings $\mathbf{P}_{T}^{\rightarrow}$ and $\mathbf{P}_{T}^{\leftarrow}$. We then adopt a transition alignment scheme to match progression representations with their corresponding transition embeddings. For each progression representation, its paired description is treated as the positive, while all other transition descriptions—including reversed-direction descriptions—serve as negatives. Similar to Eq.~(\ref{state}), the transition alignment loss is defined as
$\mathcal{L}_{\mathrm{trans}}^{\mathrm{MoD}}
= \alpha\mathcal{L}_{\mathrm{trans}} + (1-\alpha)\mathcal{L}^{\mathrm{soft}}_{\mathrm{trans}}.$

\textbf{Bidirectional Reconstruction.} While the above alignment objectives ground progression representations in the semantic space, they do not explicitly enforce temporal consistency between the learned transitions and the underlying disease states. To further regularize the progression modeling process, we introduce a bidirectional reconstruction objective that requires progression representations to faithfully describe state evolution across time. Specifically, given $\tilde{\mathbf{P}}_{I}^{\rightarrow}$ and $\tilde{\mathbf{P}}_{I}^{\leftarrow}$, we employ a lightweight reconstruction decoder $f_R$ to reconstruct target state embedding at each timepoint from the opposite state and corresponding progression representation, as $
\hat{\mathbf{H}}_{I}^{cur} = f_R([\tilde{\mathbf{H}}_{I}^{pri};\tilde{\mathbf{P}}_{I}^{\rightarrow}]),
\hat{\mathbf{H}}_{I}^{pri} = f_R([\tilde{\mathbf{H}}_{I}^{cur};\tilde{\mathbf{P}}_{I}^{\leftarrow}])
$,
where $\hat{\mathbf{H}}_{I}^{cur}$ and $\hat{\mathbf{H}}_{I}^{pri}$ denote reconstructed current and prior embeddings respectively. To enforce fine-grained consistency between original and reconstructed features, we adopt an InfoNCE-style token-level reconstruction objective. For each reconstructed token $\hat{\mathbf{H}}_{I,i}^{t,n}$ at spatial position $n$ in sample $i$, its original counterpart $\tilde{\mathbf{H}}_{I,i}^{t,n}$ is treated as the positive, while all other original tokens serve as negatives. The reconstruction loss is defined as:
{\small
\begin{equation}
\mathcal{L}_{\mathrm{recon}}
=
-\frac{1}{BN_I}\sum_{t \in \{pri, cur\}}
\sum_{i=1}^{B}\sum_{n=1}^{N_I}
\log
\frac{
\exp(\cos(\hat{\mathbf{H}}_{I,i}^{t,n},\tilde{\mathbf{H}}_{I,i}^{t,n})/\tau)
}{
\sum\limits_{j=1}^{B}\sum\limits_{m=1}^{N_I}
\exp(\cos(\hat{\mathbf{H}}_{I,i}^{t,n},\tilde{\mathbf{H}}_{I,j}^{t,m})/\tau)
},
\end{equation}}
where $\cos(\cdot,\cdot)$ denotes cosine similarity.

\textbf{Overall Objective.}
The overall optimization objective of ProTrans is $
\mathcal{L}
=
\mathcal{L}_{\mathrm{state}}^{\mathrm{MoD}}
+
\mathcal{L}_{\mathrm{trans}}^{\mathrm{MoD}}
+
\lambda_{r}\mathcal{L}_{\mathrm{recon}},
\label{total}
$
where $\lambda_{r}$ trades off the reconstruction constraint.

\section{Experiments and Results}

\textbf{Pretraining Dataset.} We constructed the pretraining dataset using MIMIC-CXR-JPG \cite{johnson2019mimic} and Chest ImaGenome dataset \cite{wu2021chest}, collecting longitudinal CXR images paired with corresponding radiology reports from MIMIC-CXR-JPG and disease progression annotations between consecutive studies from Chest ImaGenome. In total, we utilized 98,940 bi-temporal image–report pairs for VLP, excluding samples that overlapped with CXR-MS-T dataset \cite{bannur2023mscxr,bannur2023learning}. 

\noindent\textbf{Implementation Details}
We adopted ViT-B/16 as the visual encoder and BioClinicalBERT \cite{alsentzer2019publicly} as the text encoder \cite{yang2024unlocking}. AdamW was used with an initial learning rate $5\times10^{-5}$ and a weight decay 0.05. The temperature $\tau$ was set to 0.07, the momentum  0.995, trade-off coefficients $\alpha$ 0.8, and $\lambda_r$ 0.5. The progression encoder $f_P$ consists of 3 blocks, each including both spatial and temporal attention layers with 12 attention heads and an embedding dimension of 768.

\subsection{Downstream Adaptation Setup}
To evaluate the effectiveness of the learned progression representations, we performed experiments with the following two downstream tasks. 

\textbf{Task1: Disease Progression Classification.} We conducted this task on the MS-CXR-T dataset, which comprises 1,326 paired CXR images covering five findings—Consolidation, Edema, Pleural Effusion, Pneumonia, and Pneumothorax—each annotated with a three-way progression label (i.e., improving, stable or worsening). Following \cite{yang2024unlocking}, we used the visual backbone to extract progression representations $\tilde{\mathbf{P}}_{I}^{\rightarrow}$ and trained a SVM \cite{boser1992training} classifier for prediction under a 10-fold cross-validation protocol. This evaluated the discriminative ability of the learned representations without additional backbone fine-tuning. 

To further assess cross-modal semantic alignment, we adopted a classification strategy with text prototyping. Specifically, each progression category is formulated as a text prototype in the form of ``[finding] [progression]", which is encoded into text embeddings. Classification was performed by computing similarities between the progression representation and prototype embeddings, assigning the label with the highest similarity score.

\textbf{Task2: Progression Captioning.} This task aims to generate natural language descriptions that characterize changes across CXR pairs. We evaluated it on the ICG dataset \cite{ma2025towards}, which contains 11,439 paired CXRs with the corresponding radiological descriptions of visual differences. The learned progression representations were fed into a frozen medical large language model (LLM), BioMistral-7B\cite{labrak2024biomistral}, to generate free-text progression descriptions. We utilized the official splits with 10,679 samples for fine-tuning and 760 for testing. Performance was evaluated using both lexical (ROUGE-L and BERTScore) and clinical metrics (RadGraph-F1, CheXbert, GREEN and Temporal-F1) \cite{xu-etal-2025-radeval}.

\subsection{Results}
\begin{figure}[h]
    \centering
\includegraphics[width=0.65\linewidth]{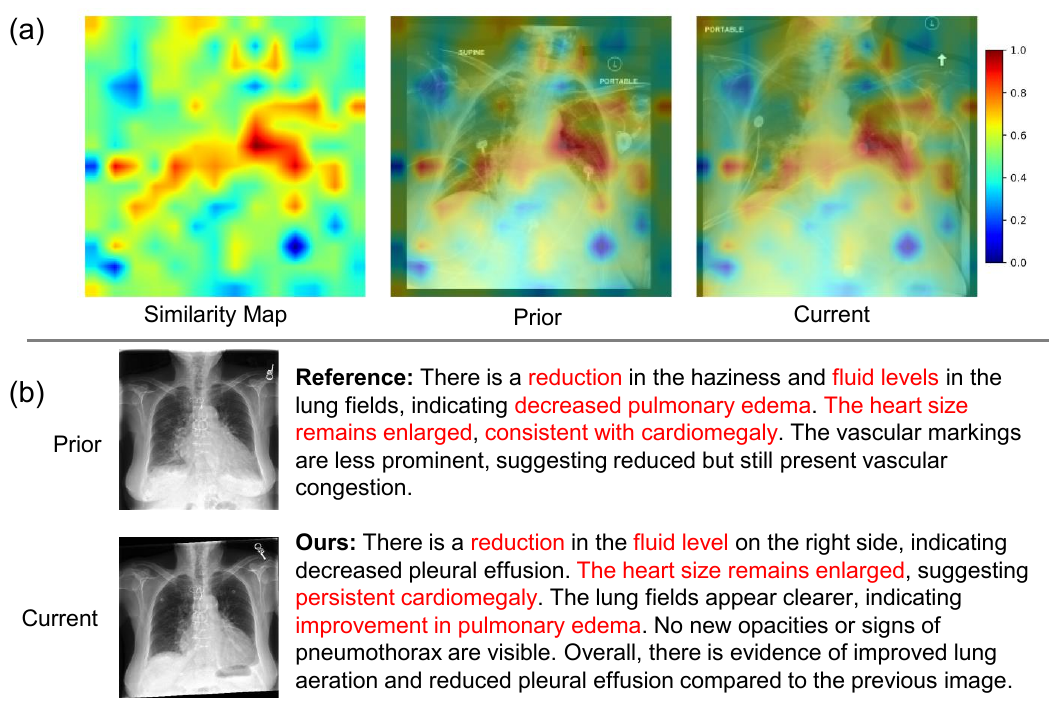}
\vspace{-5pt}
    \caption{(a) Similarity map between the learned progression features and the text prototype “edema improved”, which is projected back onto the prior and current CXR images, and (b) qualitative comparison of generated progression findings.} 
    \label{visual}
\end{figure}

\textbf{Results on disease progression classification.} As shown in Table \ref{task1_table}, our method consistently outperforms prior feature fusion-based methods (BioViL-T, MedST, TempA-VLP, Coca-CXR) and the discrepancy-based method (Diff-RRG). Using the SVM classifier, our method obtains the highest average accuracy of 63.54\%, indicating that the learned progression representations exhibit stronger linear separability. This improvement can be attributed to the explicit modeling of directional progression in the semantic space, where tighter alignment is enforced between visual progression features and direction-aware progression descriptions. As a result, the advantage of our method becomes even more pronounced in the text prototype setting. To gain deeper insights into spatial semantic correspondence, we visualize similarity maps between the learned progression representations and their associated text prototypes. As shown in Fig.\ref{visual}(a), similarity responses are mainly concentrated in lung regions. When projected back onto the prior and current CXR images, these highlighted areas correspond to clinically relevant locations associated with edema, indicating that the model effectively localizes progression-related regions.
\begin{table}[t]
\centering

\setlength{\tabcolsep}{4pt}
\centering
\small
\caption{Disease progression classification results (accuracy, \%) on the MS-CXR-T dataset. \textcolor{gray}{Gray} indicates the results cited from the original papers as the implementations are not publicly available. We calculated mean$\pm$std. over 3 random seeds. The best results are in bold. $^{\dagger}$Coca-CXR used next token prediction for classification.}
\vspace{-0.2cm}
\label{task1_table}
\small
\resizebox{\textwidth}{!} {
\begin{tabular}{lcccccc}
\toprule
Method & Consolidation & Edema & Pl.\ Effusion & Pneumonia & Pneumothorax & Avg. \\
\midrule
\multicolumn{7}{l}{\textit{SVM}} \\
BioViL-T\cite{bannur2023learning}
& 56.93{\scriptsize$\pm$1.77} 
& 61.55{\scriptsize$\pm$0.95}
& 53.94{\scriptsize$\pm$0.89}
& 67.24{\scriptsize$\pm$0.20}
& \textbf{55.46{\scriptsize$\pm$0.01}}
& 59.02{\scriptsize$\pm$0.34} \\

MedST\cite{yang2024unlocking}
& 60.57{\scriptsize$\pm$1.18}
& 67.35{\scriptsize$\pm$0.32}
& 58.47{\scriptsize$\pm$1.50}
& 65.00{\scriptsize$\pm$0.34}
& 54.18{\scriptsize$\pm$0.81}
& 61.12{\scriptsize$\pm$0.34} \\

Diff-RRG \cite{yun2025diff}
& 47.94{\scriptsize$\pm$2.71}
& 50.5{\scriptsize$\pm$2.16}
& 48.66{\scriptsize$\pm$0.00}
& 67.10{\scriptsize$\pm$0.00}
& \textbf{55.46{\scriptsize$\pm$0.01}}
& 53.93{\scriptsize$\pm$0.77} \\

\textbf{Ours}
& \textbf{61.54{\scriptsize$\pm$0.23}}
& \textbf{69.53{\scriptsize$\pm$0.64}}
& \textbf{61.81{\scriptsize$\pm$0.39}}
& \textbf{69.35{\scriptsize$\pm$0.84}}
& \textbf{55.46{\scriptsize$\pm$0.01}}
& \textbf{63.54{\scriptsize$\pm$0.29}} \\

\midrule
\multicolumn{7}{l}{\textit{Text prototype}} \\
BioViL-T\cite{bannur2023learning}
& 51.99{\scriptsize$\pm$1.21}
& 57.39{\scriptsize$\pm$0.77}
& 50.24{\scriptsize$\pm$0.00}
& 66.39{\scriptsize$\pm$0.34}
& 55.50{\scriptsize$\pm$0.22}
& 56.30{\scriptsize$\pm$0.43} \\

MedST\cite{yang2024unlocking}
& 61.06{\scriptsize$\pm$0.24}
& 65.29{\scriptsize$\pm$0.35}
& 63.03{\scriptsize$\pm$0.30}
& 67.79{\scriptsize$\pm$0.52}
& 56.14{\scriptsize$\pm$1.33}
& 62.66{\scriptsize$\pm$0.42} \\

Diff-RRG\cite{yun2025diff}
& 55.15{\scriptsize$\pm$0.74}
& 51.12{\scriptsize$\pm$0.00}
& 48.79{\scriptsize$\pm$0.73}
& 61.04{\scriptsize$\pm$0.21}
& 50.71{\scriptsize$\pm$0.71}
& 53.36{\scriptsize$\pm$0.39} \\

{\color{gray} TempA-VLP \cite{yang2025tempa}} & {\color{gray} 65.2} & {\color{gray} 77.1} & {\color{gray} 59.4} & {\color{gray} 73.4} & {\color{gray} 43.1} & {\color{gray} 64.8} \\

{\color{gray} Coca-CXR$^{\dagger}$ \cite{chen2025coca}}  & {\color{gray} 69.6} & {\color{gray} 71.8} & {\color{gray} 68.1} & {\color{gray} 56.4} & {\color{gray} 59.3} & {\color{gray} 65.0} \\

\textbf{Ours}
& \textbf{63.86{\scriptsize$\pm$0.70}}
& \textbf{69.68{\scriptsize$\pm$0.77}}
& \textbf{66.42{\scriptsize$\pm$0.11}}
& \textbf{69.05{\scriptsize$\pm$0.79}}
& \textbf{60.06{\scriptsize$\pm$0.80}}
& \textbf{65.81{\scriptsize$\pm$0.38}} \\



\bottomrule
\end{tabular}}
\end{table}

\begin{table}[t]
\centering
\caption{Progression captioning results on the ICG dataset.}
 \vspace{-0.2cm}
\label{task2_table}
\resizebox{0.8\textwidth}{!} {
\begin{tabular}{lcccccc}
\toprule
Method & ROUGE-L & BERTScore & CheXbert & Green & Radgraph-F1 & Temporal-F1 \\
\midrule
LLava-Med \cite{li2023llava} & 0.167 & 0.381 & 0.166 & 0.031 & 0.053 & 0.119 \\
LLava-Rad\cite{chaves2024towards} & 0.184 & 0.442 & 0.383 & 0.125 & 0.087 & 0.136 \\
Maira\cite{bannur2024maira}    & 0.167 & 0.415 & 0.075 & 0.077 & 0.062 & 0.118 \\
Libra \cite{zhang2025libra}    & 0.190 & 0.428 & 0.366 & 0.141 & 0.081 & 0.145 \\
Diff-RRG \cite{yun2025diff} & 0.153 & 0.407 & 0.221 & 0.101 & 0.059 & 0.110 \\

\midrule
Ours      & \textbf{0.259} & \textbf{0.535} & \textbf{0.386} & \textbf{0.230} & \textbf{0.126} & \textbf{0.238} \\
\bottomrule
\end{tabular}}
\end{table}

\noindent\textbf{Results on Progression captioning.} Table \ref{task2_table} presents progression captioning results on the ICG dataset, compared with several competing medical LLMs. Our method achieves the best performance across both lexical and clinical evaluation metrics, with large margins, about 39\% in Green and Temporal-F1, demonstrating stronger clinical fidelity and temporal awareness. Although LLaVA-Med, LLaVA-Rad, and Maira can process multi-image inputs, they are not explicitly designed to model longitudinal
relationships, thus struggling to capture subtle bi-temporal differences. Libra introduces a dedicated module to fuse multi-scale contextual information across images while Diff-RRG focuses on patch-level difference modeling. However, both methods mainly capture low-level visual variations and fail to establish semantic grounding with progression. In contrast, our method learns progression representations that encode semantic transitions between clinical states, thus providing more informative temporal cues to the language model. Fig. \ref{visual}(b) illustrates that our method accurately captures and describes clinically meaningful differences between bi-temporal images. 
\begin{table}[htbp]
\centering
\caption{Ablation study on different components of our proposed method.}
\vspace{-0.2cm}
\label{tab:ablation}
\resizebox{0.9\textwidth}{!} {
\begin{tabular}{lcccccc}
\toprule
Method & Consolidation & Edema & Pl.\ Effusion & Pneumonia & Pneumothorax & Avg. \\
\midrule

\multicolumn{7}{l}{\textit{SVM}} \\
w/o $\mathcal{L}^\text{MoD}_{state}$
& 56.41{\scriptsize$\pm$0.21}
& 62.56{\scriptsize$\pm$0.36}
& 61.15{\scriptsize$\pm$0.09}
& 69.21{\scriptsize$\pm$0.89}
& 53.87{\scriptsize$\pm$0.21}
& 60.64{\scriptsize$\pm$0.00} \\

w/o $\mathcal{L}^\text{MoD}_{trans}$
& 44.13{\scriptsize$\pm$2.04}
& 52.89{\scriptsize$\pm$0.65}
& 48.66{\scriptsize$\pm$0.00}
& 67.10{\scriptsize$\pm$0.00}
& \textbf{55.46{\scriptsize$\pm$0.01}}
& 53.65{\scriptsize$\pm$0.54} \\

w/o $\mathcal{L}_{recon}$
& 59.43{\scriptsize$\pm$1.27}
& 64.80{\scriptsize$\pm$2.02}
& \textbf{63.09{\scriptsize$\pm$0.30}}
& 66.82{\scriptsize$\pm$0.51}
& 54.36{\scriptsize$\pm$0.21}
& 61.70{\scriptsize$\pm$0.46} \\


 w/o bidirection
& \textbf{62.69{\scriptsize$\pm$0.71}}
& 66.15{\scriptsize$\pm$0.27}
& 59.20{\scriptsize$\pm$0.45}
& 66.41{\scriptsize$\pm$0.98}
&\textbf{55.46{\scriptsize$\pm$0.01}}
& 61.98{\scriptsize$\pm$0.15} \\

Ours
& 61.54{\scriptsize$\pm$0.23}
& \textbf{69.53{\scriptsize$\pm$0.64}}
& 61.81{\scriptsize$\pm$0.39}
& \textbf{69.35{\scriptsize$\pm$0.84}}
& \textbf{55.46{\scriptsize$\pm$0.01}}
& \textbf{63.54{\scriptsize$\pm$0.29}} \\

\midrule
\multicolumn{7}{l}{\textit{Text prototype}} \\
w/o pretraining
& 43.91{\scriptsize$\pm$0.25}
& 69.62{\scriptsize$\pm$1.00}
& 64.4{\scriptsize$\pm$0.64}
& 68.02{\scriptsize$\pm$1.52}
& 55.19{\scriptsize$\pm$0.00}
& 60.23{\scriptsize$\pm$0.32} \\

w/ pretraining
& \textbf{63.86{\scriptsize$\pm$0.70}}
& \textbf{69.68{\scriptsize$\pm$0.77}}
& \textbf{66.42{\scriptsize$\pm$0.11}}
& \textbf{69.05{\scriptsize$\pm$0.79}}
& \textbf{60.06{\scriptsize$\pm$0.80}}
& \textbf{65.81{\scriptsize$\pm$0.38}} \\

\bottomrule
\end{tabular}}
\end{table}

\noindent\textbf{Ablation Study.} Table \ref{tab:ablation} presents the ablation results of our framework. Removing any key module signficantly compromises the overall model performance, demonstrating their complementary roles in learning effective progression representations. In particular, eliminating the transition alignment loss $\lambda^\text{MoD}_{trans}$ leads to the most significant drop in performance, suggesting that providing explicit semantic supervision for visual progression feature learning is critical for accurately modeling longitudinal changes. Furthermore, when the progression pretraining stage is removed and the downstream text prototype model is trained directly, performance consistently deteriorates. This degradation underscores the importance of the proposed pretraining stage in providing a strong temporal semantic foundation for longitudinal downstream tasks.
\vspace{-0.3cm}
\section{Conclusion}
In this paper, we have proposed ProTrans, a novel VLP framework for progression representation learning that formulates disease progression as a directional semantic transition between bi-temporal clinical states. By leveraging radiology reports to guide state grounding and transition alignment, together with reversed temporal learning to enforce trajectory consistency, ProTrans effectively captures direction-aware and semantically grounded progression patterns. Extensive experiments on progression classification and captioning tasks demonstrate the potential of our method in learning robust progression semantics for CXRs.


%
\bibliographystyle{splncs04}
\bibliography{mybib}

\end{document}